# Good Books are Complex Matters: Gauging Complexity Profiles Across Diverse Categories of Perceived Literary Quality

**Yuri Bizzoni**
Center for Humanities Computing
Aarhus University, Denmark
yuri.bizzoni@cc.au.dk

**Pascale Feldkamp**
Center for Humanities Computing
Aarhus University, Denmark
pascale.moreira@cc.au.dk

**Mia Jacobsen**
Center Humanities Computing
Aarhus University, Denmark
miaj@cas.au.dk

**Mads Rosendahl Thomsen**
Communication and Culture
Aarhus University, Denmark
madsrt@cc.au.dk

**Kristoffer Nielbo**
Center Humanities Computing
Aarhus University, Denmark
kln@cas.au.dk

## Abstract

In this study, we employ a classification approach to show that different categories of literary "quality" display unique linguistic profiles, leveraging a corpus that encompasses titles from the Norton Anthology, Penguin Classics series, and the Open Syllabus project, contrasted against contemporary bestsellers, Nobel prize winners and recipients of prestigious literary awards. Our analysis reveals that canonical and so called high-brow texts exhibit distinct textual features when compared to other quality categories such as bestsellers and popular titles as well as to control groups, likely responding to distinct (but not mutually exclusive) models of quality. We apply a classic machine learning approach, namely Random Forest, to distinguish quality novels from "control groups", achieving up to 77% F1 scores in differentiating between the categories. We find that quality category tend to be easier to distinguish from control groups than from other quality categories, suggesting than literary quality features might be distinguishable but shared through quality proxies.

## 1 Introduction

The definition of literary "quality" has long been a subject of debate among scholars, critics, and readers alike. Expert-based quality judgments, such as literary awards, are often set in contraposition to signs of popular appreciation, observed for example in what appears on bestseller lists or has high ratings online (Algee-Hewitt et al., 2016; Porter, 2018; Underwood and Sellers, 2016). An often discussed dimension of literary quality is that of the so-called "literary canon", a complex concept generally denoting a set of works that have survived in the memory of a literary culture (Bloom, 1995). As a collective process of cultural selection, no one individual authority bestows (and can point to the features of) canonicity, which makes the very definition of the canon complex. Canonical literature can be considered a mid-way entity: it is the result of the fine-grained selections of large amounts of people over time, but it is also "curated", disseminated and validated by literary elites (Shesgreen, 2009). Some schools of literary scholars – most notably one side of the "canon-wars" of the canonicity-debate of the 1980s – have held the canon to represent nothing but entrenched interests (von Hallberg, 1983), or the cultural capital of current ruling classes (Guillory, 1995), while others have maintained that "canonic" works excel in terms of some set of intrinsic textual features, though vaguely defined (Bloom, 1995; van Peer, 2008). The quantifiable characteristics that distinguish canonic from non-canonic works, but also from other categories of "literary quality", like best-sellers or prestigious award-winning books, if any, remain elusive, or are framed in vague and undefined terms (*powerful prose*, *great humor*, *smooth development*, etc.). This study seeks to bridge this gap by employing computational techniques to explore the linguistic profiles that differentiate these nuanced categories of literary prestige.

While computational linguistics has made significant strides in text analysis, its application to literary studies has predominantly focused on authorship attribution or genre classification. Moreover, it has often revolved around modelling what can be broadly labelled stylistic features as bags-of-words (Da, 2019; Bode, 2023). There is a notable gap in research that utilizes these and more sophisticated sets of textual and narrative features to investigate literary quality.Specifically, the comparative analysis of "literary quality" as a mutlifaceted category – including canonical works, prestigious award-winning novels, and bestsellers against control groups – in terms of linguistic attributes has been underexplored.

In this work we leverage a large corpus that spans various categories of "quality", by including bags-of-words stylistic measures as well as linear features of narrative complexity, to study the linguistic profiles of different categories of literary quality, providing empirical evidence to informs the ongoing discourse on literary prestige and merit.

The overarching hypotheses of this paper are:

- **H1**: There are linguistic/textual features that distinguish "high quality" literature; and

- **H2**: Their profiles change depending on what kind of quality category or "perception of quality" we take into consideration.

## 2 Related works

There have been many rules and recommendations about how to write better, supposedly applicable across genres and to both high and low-brow literature, from detail-oriented suggestions about which parts of speech one ought to avoid to funny rituals inducive to writing. Sherman (1893) proposed that simplicity – i.e. shorter sentences – should be a marker of a "better" style. Readability indices have in this regard been thought to hint not only at the accessibility of a text, but implicitly at its "quality", and are widely implemented in more recent creative writing and publishing aids.[1]. Still, the importance of the readability of a literary text in the context of reader appreciation is controversial (Martin, 1996; Garthwaite, 2014). Studies seeking to predict literary success or perceived quality do, however, follow the intuitive idea that readers perceive a difference between "difficult" and "easy" fiction, tending to approximate some form of stylistic complexity by using textual features related to readability (i.a., sentence-length, vocabulary richness, redundancy)(Brottrager et al., 2022; van Cranenburgh and Bod, 2017; Crosbie et al., 2013; Koolen et al., 2020; Maharjan et al., 2017; Algee-Hewitt et al., 2016). In general, the focus on some form of **literary complexity** is not new in western culture. Some "simplicity laws" for literature have traditionally been set forth by critics and writers alike – for example, Hemingway recommends a direct and personal style (Hemingway, 1999). A highly popular if not canonic author, King, advocates more readable texts King (2010); and Strunk et al. (1999)'s influential literary theory book, *The Elements of Style*, advised, i.a., using the active voice and avoiding redundancy. Conversely, others have promoted "purple prose",[2] characterized as a complex and challenging style, "rich, succulent and full of novelty" (West, 1985). However contradictory these positions may seem, both may hold merit for different ways of understanding literature. Regarding the "difficulty" of prose, at least in terms of readability, reader preference appears to be audience-specific (Bizzoni et al., 2023a). In examining the canon, computational literary studies have predominantly followed the same line of modelling stylistic features (Brottrager et al., 2022; Barré et al., 2023). Often, the profile of canonic works has been connected to some form of textual complexity, whether in the form of lower readability (Bizzoni et al., 2023a), textual entropy (Algee-Hewitt et al., 2016) or higher perplexity and cognitive demand on the reader (Bizzoni et al., 2023c). Moreover, features of style are seen to vary across "types" of literature: award-winning works are *less readable*, while *more readable* books appear to score higher on GoodReads (Bizzoni et al., 2023a). Similarly, more prestigious literature appears to elicit higher perplexity (i.e., LLM perplexity) than popular literature (Wu et al., 2024). Computational studies seeking to model reader appreciation and/or canonicity have predominantly focused on **the stylistic level**, modelling distributions of stylistic features in bag-of-words or bag-of-sentences approaches, ranging from the most basic measures of difficulty or complexity, such as sentence length (Maharjan et al., 2017; Mohseni et al., 2022), to more experimental measures like compressibility of a text file, aiming to identify stylistic signatures or markers of literary quality (Archer and Jockers, 2017; Koolen et al., 2020; Wang et al., 2019). Subsequent research expanded into sentiment analysis (SA), examining how emotional dynamics within a narrative – intensity, fluctuations, and trajectory – can influence reader perception and engagement. Much of this work has centered on tracing so-called sentiment arcs, i.e., time-series resulting from sequentially scored words or sentences with SA methods (Jockers, 2014). This focus on the emotional landscape of texts introduced a novel lens for understanding narrative techniques and their impact on the reader experience (Hogan, 2011; Cambria et al., 2017), with potential for moving beyond the stylistic level in modelling perceptions of literary quality (Pianzola et al., 2023). Still, questions persist as to how to operationalize an affective narratology (Rebora, 2023) – that is, for example, are sentiment arcs of novels as derived through SA tools actually palpable to readers? While most studies have focused on the visual shapes of sentiment arcs (Reagan et al., 2016; Jockers, 2015), others have applied more sophisticated measures to gauge their shapes and approximate complexity at **the narrative level** (Maharjan et al., 2018; Bizzoni et al., 2022), on the intuition that readers tend to appreciate certain shapes, or a certain balance in the complexity of narrative flow. Hu et al. (2020) and Bizzoni et al. (2022) have modeled the persistence, coherence, and predictability of arcs through measures like the Hurst coefficient and Approximate Entropy (ApEn) to measure global and local complexity (Bizzoni et al., 2023b). Such measures appear to be applicable for distinguishing between types of literary prestige (Bizzoni et al., 2021, 2023c).This perspective aligns with theories that emphasize the narrative's capacity to engage and challenge readers, proposing narrative or sentiment complexity as a key determinant of literary quality (Hu et al., 2020). Moreover, it draws on studies observing the role of fractal patterns or entropy for aesthetic attraction (Cordeiro et al., 2015; McGavin, 1997) also in other domains, such as in music or the visual arts (McDonough and Herczyński, 2023; Brachmann and Redies, 2017).

---

[1]Such as the Hemingway, or Marlowe applications

[2]A notion derived from Horace's *Ars Poetica*; in which "weighty openings and grand declarations" are said to "have one or two *purple patches* tacked on, that gleam / far and wide" (Horace, 2005).

## 3 Methods

Drawing on the insights from literary theory and computational study on features and profiles of literary quality, we focus on narrative complexity in modelling the feature profiles of various categories of perceived literary quality. Our approach not only adopts a multi-level perspective on literary quality itself, but also on literary complexity, examining complexity at the stylistic and syntactic level (including simple features, such as vocabulary richness and deeper features, such as perplexity) and at the level of sentiment or narrative (including simple features, such as mean valence, and deeper features, such as sentiment arc entropy) across a diverse corpus of literary works.

### 3.1 Corpus

Our corpus comprises a carefully curated selection of 9,089 novels of various genres, published in the US between 1880 and 2000 (see Table 1 and Figure 1). It is a unique dataset both in terms of size [3] and diversity, as the corpus was compiled based on the number of libraries holding each novel, with a preference for more circulated works. Library holdings reflect a diverse demand, therefore the corpus is not homogeneous in terms of genre and lists both prestigious and popular works ranging from Nobel prize winners to Science Fiction classics (Long and Roland, 2016).[4]

---

[3]Often, studies on reader appreciation rely on < 1,000 books (Ganjigunte Ashok et al., 2013; Koolen et al., 2020).

[4]The corpus has no reference publication, though other studies are based on it (Underwood et al., 2018; Cheng, 2020). See https://textual-optics-lab.uchicago.edu/us_novel_corpus for a corpus description.

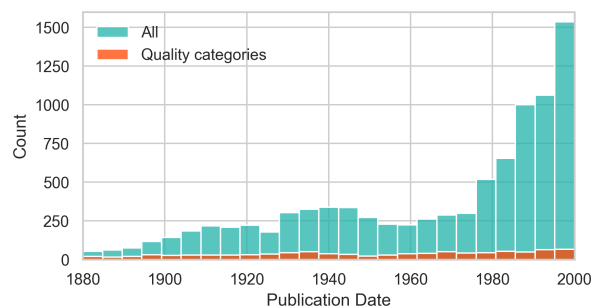

Figure 1: Distribution of titles in categories of perceived quality (Canon, Awards, Nobel, and Bestseller groups) in the Chicago Corpus over time.

| Category | Titles | Authors | Titles/Author |
|---|---|---|---|
| All | 9089 | 3166 | 2.87 |
| Canon | 618 | 163 | 3.80 |
| Nobel | 85 | 18 | 4.72 |
| Prizes | 144 | 108 | 1.33 |
| Bestsellers | 228 | 130 | 1.75 |
| Rest | 7955 | 2933 | 2.71 |

Table 1: Number of titles, authors, and average titles per author in the dataset and for each quality category. Note that "Rest" denotes titles that are included in neither quality category.

### 3.1.1 Quality categories

We divided titles into different categories of perceived quality (Table 1). We considered novels that bear some mark of perceived quality those that: (i) are canonic in the sense that the they often appear on college syllabi,[5] are included in the most prominent literary anthology,[6] or in a publisher's classics series;[7] (ii) are by Nobel prize-winning authors; (iii) have been long-listed for prestigious literary awards;[8] (iv) are listed on bestseller lists of the 19th and 20th century.[9] The amount of works that fall inside one of these categories is relatively consistent across decades (Fig. 1).

It should be noted that we have sought to make the classification among what we refer to as different – though overlapping – markers of quality difficult. The novels in quality categories do not necessarily stand out in terms of stylistic and narrative quality from those not selected. For example, the corpus contains important works of genre-fiction (i.e., Tolkien or Philip K. Dick) as well as influential authors of popular fiction (such as Agatha Christie and Stephen King). It should be noted that the presence of such other classics and popular titles that do not fall within any of the mentioned categories increases the difficulty of a classifier's tasks. Naturally, we consider the division into categories an artificial, though necessary heuristic to make the study possible. In fact, canonicity is neither defined nor boolean (Barré et al., 2023), but may be best represented as a continuum on several dimensions. Similarly, the overlap of the categories should increase the difficulty of differentiation (see Fig. 2) as in numerous cases a text might be, for example, both canonical and a bestseller. In some sense we challenging the classifier to see whether these categories are representative of distinctive profiles - of which a novel can contain more than one, as it ends up in more than one category.

### 3.1.2 High/low GoodReads ratings

Beyond the categories of perceived quality, we collected the highest and lowest rated titles on GoodReads, a large online platform for rating and reviewing books. With its 90 million users, GoodReads arguably offers an insight into reading culture "in the wild", cataloguing books from a wide spectrum of genres (Nakamura, 2013). It derives book-ratings from a heterogeneous pool of readers in terms of background, gender, age, native language and reading preferences (Kousha et al., 2017). We distinguished classes at 3.8 average GoodReads rating,[10] where we consider high-rating titles those that are rated above (n=4680), and low-rating those that are rated below or equal to this threshold (n=4387).

### 3.2 Features

To capture the complexity of the literary texts at various levels, we extracted a set of stylistic and narrative features that both approximate some form of complexity and have been known to influence perceptions of literary quality. A description of each feature including reference studies are listed in Table 2. We divide these features into **stylistic features** (with a subcategory of more syntactic features) and **narrative features**, where the former are surface features, calculated using a bag-of-words approach and the latter are higher-level features based on sentiment analysis, where the complexity measures Approximate entropy and the Hurst exponent take the progression of novels into account.

### 3.3 Model and Evaluation

We employed a "classic" machine learning model to classify novels based on the extracted features:

---

[5] We relied on the OpenSyllabus database, which indexes 18.7 million college syllabi: https://www.opensyllabus.org; tallying all works in our collection by the top 1000 most frequent authors in *English Literature* syllabi.

[6] We used the English and American edition of *the Norton Anthology*, which is often referred to as indexing canon (**??**), marking all books by authors indexed.

[7] As one of the – if not *the* – most prominent classics series (Alter et al., 2022), we used the Penguin Classics, marking titles in the corpus that are also printed as part of the series.

[8] We marked all titles extant in the corpus that were long-listed for the Pulitzer Prize and the National Book Award.

[9] Contained in the Publisher's Weekly bestseller list or the New York Times bestseller list.

[10] The threshold is justified by its mid-scale position considering the general positive skew of ratings (see the distribution of ratings in the Appendix), and as we sought to have equally sized low and high rating categories.

| Feature | Description | Type | Reference |
|---|---|---|---|
| **Type-Token Ratio** | Measures lexical diversity by comparing the variety of words (types) to the total number of words (tokens) in a text, indicating a text's vocabulary complexity and inner diversity (Torruella and Capsada, 2013).[a] | Stylistic | Forsyth (2000)*, Kao and Jurafsky (2012)*, Algee-Hewitt et al. (2016), Maharjan et al. (2017), Koolen et al. (2020), Brottrager et al. (2022), Jacobs and Kinder (2022), Bizzoni et al. (2023b) |
| **Readability** | Estimate reading difficulty based variously on sentence length, syllable count and word length/difficulty. Assessed using five different classic formulas that remain widely used (Stajner et al., 2012).[b] | Stylistic | Martin (1996), Garthwaite (2014), Maharjan et al. (2017), Febres and Jaffe (2017), Zedelius et al. (2019)*, Berger et al. (2021)*, Brottrager et al. (2022), Bizzoni et al. (2023a) |
| **Compressibility** | Measures the extent to which the text can be compressed, serving as an indirect indicator of redundancy and lexical variety (Ehret and Szmrecsanyi, 2016).[c] | Stylistic | van Cranenburgh and Bod (2017), Koolen et al. (2020), Bizzoni et al. (2023b) |
| **Passive/active ratio** | Quantifies the number of active against passive verbs in the text, associated to a better style (King, 2010). | Stylistic/ Syntactic | Hye-Knudsen et al. (2023), Wu et al. (2024) |
| **Nominal style ratio** | Quantifies the proportion of nouns and adverbs (over verbs) in the text, reflecting the nominal tendency in style, which is often associated with complex linguistic structures, denser communicative code, expert-to-expert communication (McIntosh, 1975; Bostian, 1983). | Stylistic/ Syntactic | Charney and Rayman (1989)*, Crossley et al. (2014)*, Wu et al. (2024) |
| **"Of"/"that" frequencies** | Frequency of these function words have been seen to indicate, in the case of "of", a more nominal prose, and in the case of "that", a more declarative and verb-centered prose. a more declarative or nominal style. | Stylistic/ Syntactic | Wu et al. (2024) |
| **Perplexity** | Represents the predictability of the prose through three different large language models (GPTs).[d] Higher values indicate greater complexity or unpredictability. | Stylistic/ Syntactic | Sheetz (2018), Wu (2023), Wu et al. (2024) |
| **Mean valence** | Represents the average sentiment of the text (positivity or negativity).[e] | Narrative/ Sentiment | Veleski (2020), Pianzola et al. (2020)*, Berger et al. (2021)*, Jacobs and Kinder (2022), Pianzola et al. (2023), Bizzoni et al. (2023b) |
| **Valence SD** | Represents the average variability in sentiment, indicating the range of sentiment within the narrative.[e] | Narrative/ Sentiment | Berger et al. (2021)*, Bizzoni et al. (2023b) |
| **Hurst exponent** | Quantifies the long-term auto-correlation of the sentiment arc,[e] with higher values suggesting a more complex, self-similar structure across different scales.[f] | Narrative/ Sentiment | Mohseni et al. (2021), Bizzoni et al. (2021), Bizzoni et al. (2023c) |
| **Approximate entropy** | Assesses the predictability of sequences of the sentiment arc,[e] with lower values indicating greater regularity or simplicity.[f] | Narrative/ Sentiment | Hu et al. (2020), Mohseni et al. (2022), Bizzoni et al. (2023b) |

Table 2: **Features related to stylistic and narrative complexity**. "References" refer to studies that have included the given feature and shown some relation between the feature and reader appreciation, success, or canonicity. Note that this table only includes features chosen for this study. * Denotes studies on objects connected to cultural success, however in relevant domains other than *established prose fiction* (e.g., online stories, movies).

[a] We used a common method insensitive to text length: the Mean Segmental Type-Token Ratio (MSTTR). MSTTR-100 represents the overall average of the local averages of 100-word segments of each text.
[b] Flesch Reading Ease, Flesch-Kincaid Grade Level, SMOG Readability Formula, Automated Readability Index, and New Dale–Chall Readability Formula.
[c] We calculated the compression ratio (original bit-size/compressed bit-size) for the first 1500 sentences of each text using bzip2, a standard file-compressor.
[d] All perplexity calculations were via gpt2 models, done on the byte pair encoding tokenization used in the series of gpt2 models. To get the mean perplexity per novel, we used a sliding window due to maximum input length. For details on the computation, see Wu et al. (2024).
[e] All sentiment analysis was performed using nltk's VADER implementation on a sentence-basis (compound score per sentence). For complexity measures (Hurst and ApEn, we used both VADER and the widely used Syuzhet dictionary to extract the sentiment arcs on which these measures are based.
[f] For details on the measure, please refer to Bizzoni et al. (2023c).

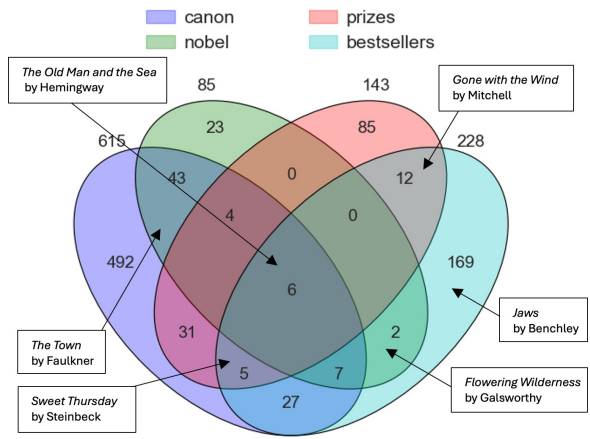

Figure 2: Number and overlap of the quality categories used in this study. The boxes give examples of titles contained in intersecting areas. Note that the largest overlap appears to be between the canon and prizes, indicating the close relation between the two. Still, in terms of percentages, the canon and Nobel categories show the largest overlap.

Random Forest (Breiman, 2001). We chose the Random Forest for its robustness to overfitting and ability to handle nonlinear relationships. In each of the following experiments we configured the classifier with 900 trees and trained on 80% of the relevant subset. Model performance was assessed using the accuracy and F1 score, enabling a balanced evaluation of both false positives and false negatives. Additionally, we conducted a feature ablation study to understand the impact of removing specific features (e.g., stylometric features, perplexity) on the classification accuracy, providing insights into the relative importance of different complexity measures in predicting literary quality.

## 4 Results

### 4.1 Performance

#### 4.1.1 Sampling

We used random subsampling for balancing the dataset. To mitigate the risk of aleatory results, in the rest of the paper all reported results will be averaged over ten independent runs, each run training and testing on a new subset where the majority class was randomly subsampled. All classifications are run on balanced classes.

#### 4.1.2 Binary classification

In binary classification tasks, we evaluated the performance of our models using different subsets of features, achieving balancing through repeated random subsampling. The variation in precision, recall, and F1 scores across different feature sets (see Fig. 5) indicates the differential predictive power of the features. The highest F1 score was achieved when all proposed features were included (Table 3), reinforcing the hypothesis that a multifaceted approach to textual analysis is crucial for accurate classification.

#### 4.1.3 Multi-class

The results of the multi-class classification task are summarized in Fig 4. The matrix reveals the model's performance in classifying texts into the five categories: canonical works, awarded works, Nobel works, bestsellers, and high/low GoodReads ratings. Notably, the model demonstrates a strong ability to distinguish awarded texts, with a substantial number of true positives. However, there is some confusion between canonical works and bestsellers, indicating areas where the feature set may not fully capture the distinguishing characteristics between these two categories.

### 4.2 Feature impact

The analysis of feature impact demonstrates the intricate nature of literary quality and the necessity of a multilevel approach to textual analysis.

Each feature contributes uniquely to the model's ability to discern among categories of literary quality, and the combined use of stylistic and narrative features enriches the classification process.

#### 4.2.1 Stylistics

The so-called stylistic features alone, including TTR, compressibility and readability scores, had a noticeable impact on all models' performance, suggesting that this level of stylistic complexity - lexical diversity and a composition of sentence length and word complexity - is a significant marker of literary quality through most considered dimensions. This category is especially useful in distinguishing canonical novels and GoodReads higher-rated books from their relative control groups, and its absence brings the bestsellers classifier to its lowest performance. As we show in Fig. 3, bestsellers exhibit a higher TTR, suggesting a wider range of vocabulary usage compared to other textual categories like long-listed novels, which – perhaps surprisingly – do not display a high TTR. Still, canonical novels predominantly tend to have a systematically higher level of readability scores, characterizing a more complex language usage, while bestsellers tend to have lower readability scores, reflecting simpler language and sentence structure, and both the award group and the Nobel group also show higher scores than the other categories. Overall, canonical texts appear to be the most demanding in terms of readability, in alignment with a previous study (Bizzoni et al., 2023a).

|  | Canon/not | Awards/not | Nobel/not | Bestseller/not | High/low GR | Multiclass |
|---|---|---|---|---|---|---|
| **N. samples** | 1236 | 288 | 170 | 456 | 8774 | 5755 |
| **F1** | .77 (.02) | .7 (.03) | .76 (.07) | .7 (.05) | .63 (.009) | .36 (.08) |
| **Accuracy** | .75 (.02) | .65 (.02) | .76 (.06) | .68 (.04) | .64 (.006) | .41 (.07) |

Table 3: F1 score and accuracy per category and for the multiclass classification. Values in parenthesis are the standard deviation. Note that GR stands for GoodReads Rating.

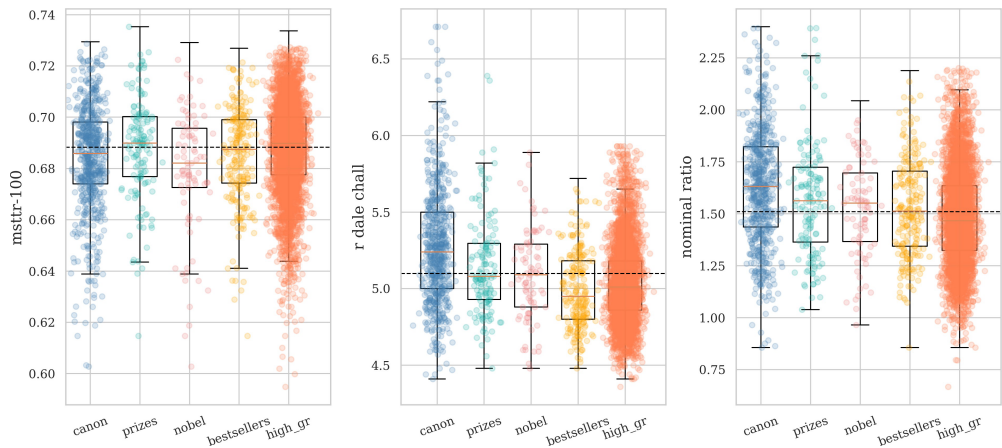

Figure 3: Boxplots indicating – from left to right – the levels of TTR, Readability, and Nominal Ratio per quality category. The black dashed line indicates the corpus mean value per feature.

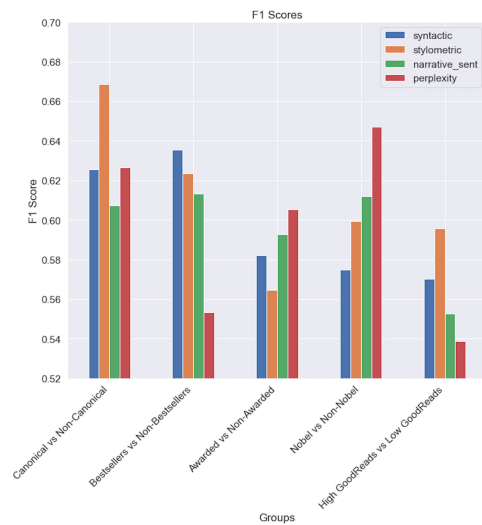

Figure 4: Confusion matrix of the multiclass experiment.

#### 4.2.2 Syntactic features

The syntactic features we selected appear very important on their own – especially in differentiating between bestsellers and non-bestsellers (Fig. 5) – and their absence harms the performance of the classifier for the awards, the bestsellers and the GoodReads categories (Tab. 4). When combined with other features, they still indicate the impor-tance of syntactic complexity also in distinguishing canonic and non-canonic literature. We found that canonical and high-brow award-listed novels make a marked use of the nominal style with respect to both bestsellers and the rest of the books in our corpus. We find this strategy surprising in literature, as it is used in other domains to optimize the communication of information in an expert-to-expert setting (Degaetano-Ortlieb and Teich, 2022). The inverse relation of nominal style with readability indices – based on sentence and word length – also seems to play within this frame.

Figure 5: Performance for each category per features set in isolation.

|                      | Canon/not | Awards/not | Nobel/not | Bestseller/not | High/low GR |
|----------------------|-----------|------------|-----------|----------------|-------------|
| - Stylistic          | .71       | .67        | .68       | .68            | .60         |
| - Perplexity         | .74       | .63        | .68       | .69            | .61         |
| - Styl./Syntactic    | .76       | .64        | .75       | .65            | .61         |
| - Narrative/Sentiment| .75       | .69        | .71       | .74            | .63         |

Table 4: F1 score per category against control for the **ablation experiment**. Each row represents the features that were *removed* before performing the classification.

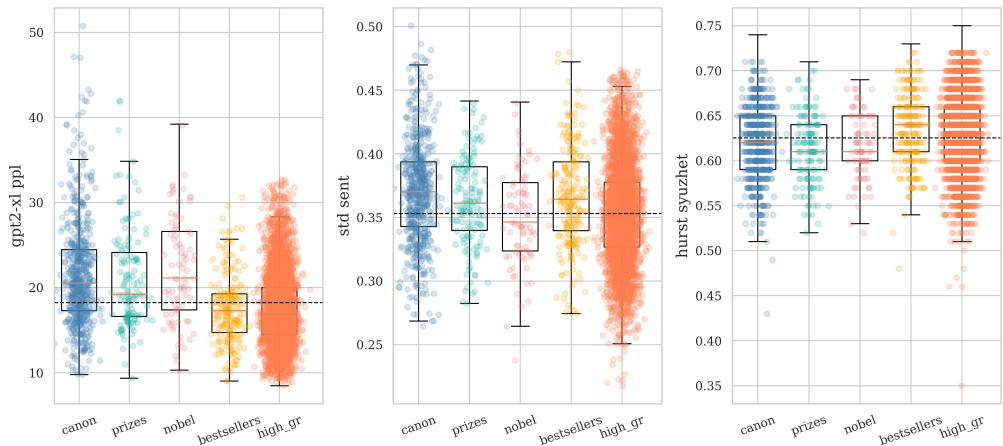

Figure 6: Boxplots indicating – from left to right – the levels of Perplexity, Valence SD, and Hurst exponent per quality category. The black dashed line indicates the corpus mean value per feature.

### 4.2.3 Perplexity

Perplexity, as a measure of the predictability of a text, represents the amount of information that is communicated by the text in terms of information theory. From an information theory perspective, a more perplexing text contains more information and is in that sense more complex or informationally "dense". We see that perplexity has a strong impact on all classifications, and especially on the differentiation of canonical, award, and Nobel groups from control-groups (Fig. 5). Perplexity appears lower than average in bestsellers and in the high GR rating group, suggesting a higher degree of predictability and simplicity in their language (Fig. 6). In contrast, canonical novels and Nobel texts show the highest perplexities, alluding to more complex language usage that requires greater cognitive effort to process (Fig. 6). This finding aligns with another recent study (Wu et al., 2024), and together with the higher nominal style of canonic texts suggests that there is a particular "canonic profile" of works, which uses language less expectedly and manages to reach a particularly high information density. A similar mechanism seems to be at work for the Nobel texts.

### 4.2.4 Narrative features

The sentiment features' predictive power has improved the performance of at least some categories. The variability of sentiment (valence SD) seems more pronounced in two usually opposed metrics of "literary quality", canonical novels and bestsellers (Fig. 6). Canonical texts have a particularly high valence SD, showcasing the ability to move frequently through a broader emotional range. The Hurst exponent is the highest for bestsellers (Fig. 6), suggesting a more self-similar and less complex narrative structure over various scales. Canonical, Nobel and long-listed texts, on the other hand, show Hurst exponents that are lower than average, indicating a higher complexity and less self-similarity in their narrative structures. The same categories of texts that use a more compact, nominal style seem able to build less self-similar narrative arcs and to range through a broader sentimental spectrum (standard deviation in sentiment scores). While these features appear to pick on a weaker signal than the others (Fig. 5), a decrease in performance is observed when they are removed from the feature set (compare with full results in Table 4), highlighting the importance of these high-level complexity metrics in capturing an aspect of the narrative structure that is not grasped by the other features.

## 5 Discussion & Conclusion

It is possible to an extent to predict the perceived quality of a text from its features (H1) and different proxies seem to be valuing different complexity features at different levels (H2).

Canonical works, bestsellers, Nobel laureates' and award-winning works, and high-rated novels on GoodReads each exhibit unique profiles with respect to the "control populations" represented by the rest of our corpus across various stylistic and narrative dimensions, and could be positioned on a multi-dimensional "complexity" continuum. At the same time, the difficulty of telling them apart in a multi-class classification experiment shows that they also represent overlapping profiles (partly explained by their de facto overlap, seen Fig. 2).

We found canonical texts to have the most distinctive profile across all dimensions and to be the easiest to classify in the binary classification task. These have a denser nominal style and lower readability scores while maintaining a less predictable sentimental line. We also found that canonical novels are more perplexing than any of the other groups, followed by award-listed high-brow novels. Such complexity, which is held to require greater cognitive effort (McIntosh, 1975), may be one contributing factor to the lasting impact and classification of these works as 'canonical'. It appears to be also partly shared by the long-listed novels and the books of Nobel laureates. In the multi-class classification, these three groups are easily confused with each other. On the other hand, bestsellers, characterized by a somewhat opposite profile, display an increased readability, lower perplexity, and a higher Hurst exponent. Together with the group of novels more highly praised on GoodReads, yet to a higher extent, they seem to employ a more accessible and predictable language, which could account for their mass appeal and commercial success. For these works, easier is better (Sherman, 1893; King, 2010). It seems that from a somewhat abstract information-theoretical perspective, canonical and awarded novels are able to apply general linguistic strategies to communicate more efficiently, at the price of a larger cognitive effort required to the reader. But the question of what exactly are these texts doing – how they are using each level of complexity – remains an open question for future studies. Finally, it is worth noting how binary classification tends to report higher results than multi-class. While this is partly to be expected from the nature of the experiment, it might also suggest that "quality profiles" are indentifiable but shared through different quality proxies, pointing to a more universal perspective on what has high quality in literary works. Future research should aim to expand the corpus, integrate more diverse (non-Anglophone) literary traditions, and explore the temporal dynamics of literary quality. Also, a better understanding of the implications of studying literature in terms of complexity is necessary to further the study of this aspect of the literary domain.

## Limitations

The selection of texts, while extensive, is not exhaustive and may reflect biases inherent in the compilation of canonical and award-winning lists. One important limitation of our corpus of novels is its strong Anglophone and American tilt: there are few non-American and non-Anglophone authors, which inevitably situates the entire analysis within the context of an Anglophone literary field.

Regarding the proxies of reader appreciation used in this study, it is hard to control the demographics of each proxy for literary quality and reception. Generally, sources like GoodReads are more diverse and represent a more comprehensive demographic selection than awards committees or anthologies' editorial boards, which are also susceptible to quick changes. Still it should be noted that the majority of GoodReads users from the beginnings of GoodReads in 2007 were native English speakers, which may affect the way users value non-Anglophone literary productions. Additionally, it is likely that there is a correlation between reviews on GoodReads and the quality categories suggested in this study, but as with any proxy measurement, it is difficult to concretely distinguish popularity, success, and quality.

Finally, the interpretation of complexity and its relation to quality is culturally and temporally situated and may change with both shifting literacy standards and literary norms.

## A  Appendix

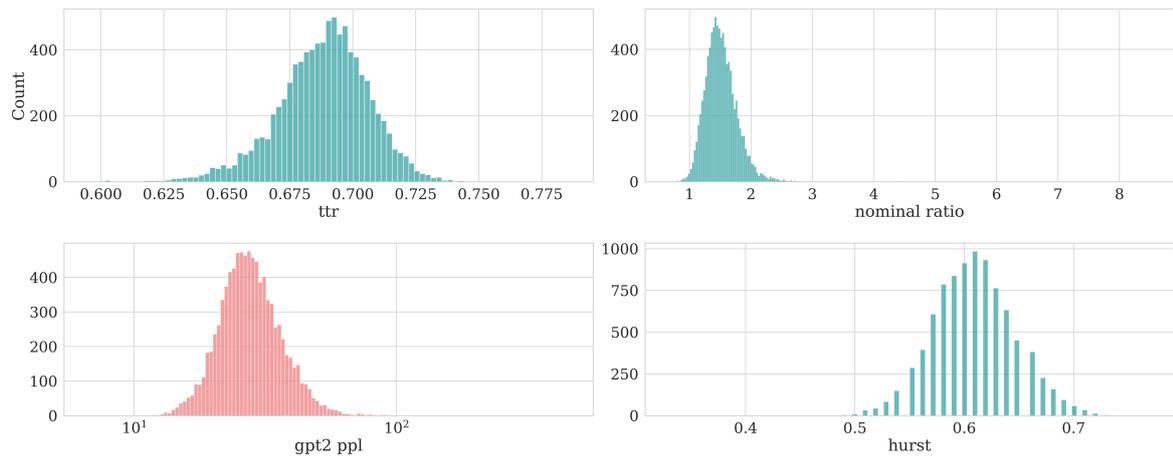

Figure 7: Distribution of selected features from each feature-type (Stylistic, Stylistic/syntactic, and narrative/sentiment).

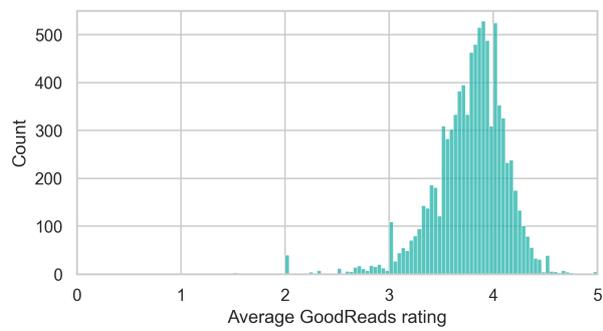

Figure 8: Distribution of average GoodReads ratings in our corpus. Note the noticeable positive skew of ratings.